\documentclass{article} 
\usepackage[preprint]{colm2026_conference}

\usepackage{amsmath,amsfonts,bm}

\def\eqref#1{equation~\ref{#1}}

\def\1{\bm{1}}

\DeclareMathAlphabet{\mathsfit}{\encodingdefault}{\sfdefault}{m}{sl}
\SetMathAlphabet{\mathsfit}{bold}{\encodingdefault}{\sfdefault}{bx}{n}

\usepackage{microtype}
\usepackage{hyperref}
\usepackage{url}
\usepackage{xcolor}
\usepackage{float}
\usepackage{amssymb}
\usepackage{amsthm}
\usepackage{amsmath}
\usepackage{amsfonts}
\usepackage{array}
\usepackage{wrapfig}
\usepackage{multirow}
\usepackage{booktabs}
\usepackage{graphicx}
\usepackage{colortbl}

\usepackage{lineno}

\definecolor{darkblue}{rgb}{0, 0, 0.5}
\hypersetup{colorlinks=true, citecolor=darkblue, linkcolor=darkblue, urlcolor=darkblue}

\title{Tuning Language Models by Mixture-of-Depths Ensemble}

\author{Haoyan Luo \\
University of Cambridge \\
\texttt{hl678@cam.ac.uk} \\
\And
Lucia Specia \\
Imperial College London \\
\texttt{l.specia@imperial.ac.uk}
}

\begin{document}

\ifcolmsubmission
\linenumbers
\fi

\maketitle

\begin{abstract}
    Transformer-based Large Language Models (LLMs) traditionally rely on final-layer loss for finetuning and final-layer representations for predictions, potentially overlooking the predictive power embedded in late layers. 
    Interpretability tools such as the logit lens show that late-layer representations already carry largely formed, task-relevant predictions; here we ask whether that observation can be turned into an actionable training signal. 
    We find that focusing tuning effort on these layers can yield losses comparable to those of the final layer, with complementary test-time behaviour. Building on this, we introduce a tuning framework, \textit{Mixture-of-Depths Ensemble} (MoDE), which treats the late layers as an ensemble that contributes to the final logits through learned routing weights. 
    MoDE can be applied on top of any existing tuning method (e.g.\ LoRA) and, in our experiments, modestly improves reasoning performance at a small parameter overhead.
    We present MoDE as a mechanism study showing that late-layer logits can be made directly useful for tuning, and that they can substitute for substantially larger trainable modules with comparable performance.
\end{abstract}

\section{Introduction}
\label{sec:intro}
    Large Language Models (LLMs) are predominantly Transformer-based, processing sequences of input tokens by representing them as vectors and transforming them through multiple layers of transformers \citep{vaswani2017attention}. Prior research has demonstrated that the intermediate representations can carry meaningful information \citep{li2024inference}, and leveraging these representations during decoding can improve trustworthiness \citep{chuang2023dola} and reasoning capabilities \citep{o2023contrastive}. While intermediate layers have been trained for inference efficiency through early exit \citep{schuster2022confident, elhoushi2024layerskip} and steered through representation finetuning \citep{wu2024reft}, comparatively little work uses their \emph{logits} as an explicit training signal that contributes to the final prediction. In the standard setup, each layer transformation creates new token representations added to the residual stream, yet only the final layer representations are used to obtain the training loss. Consequently, loss minimisation directly optimises these final representations, leaving hidden representations optimised only implicitly, thereby obscuring their potential predictive power.\looseness=-1

\begin{wrapfigure}{r}{0.5\textwidth}
        \centering
        \includegraphics[width=0.48\textwidth]{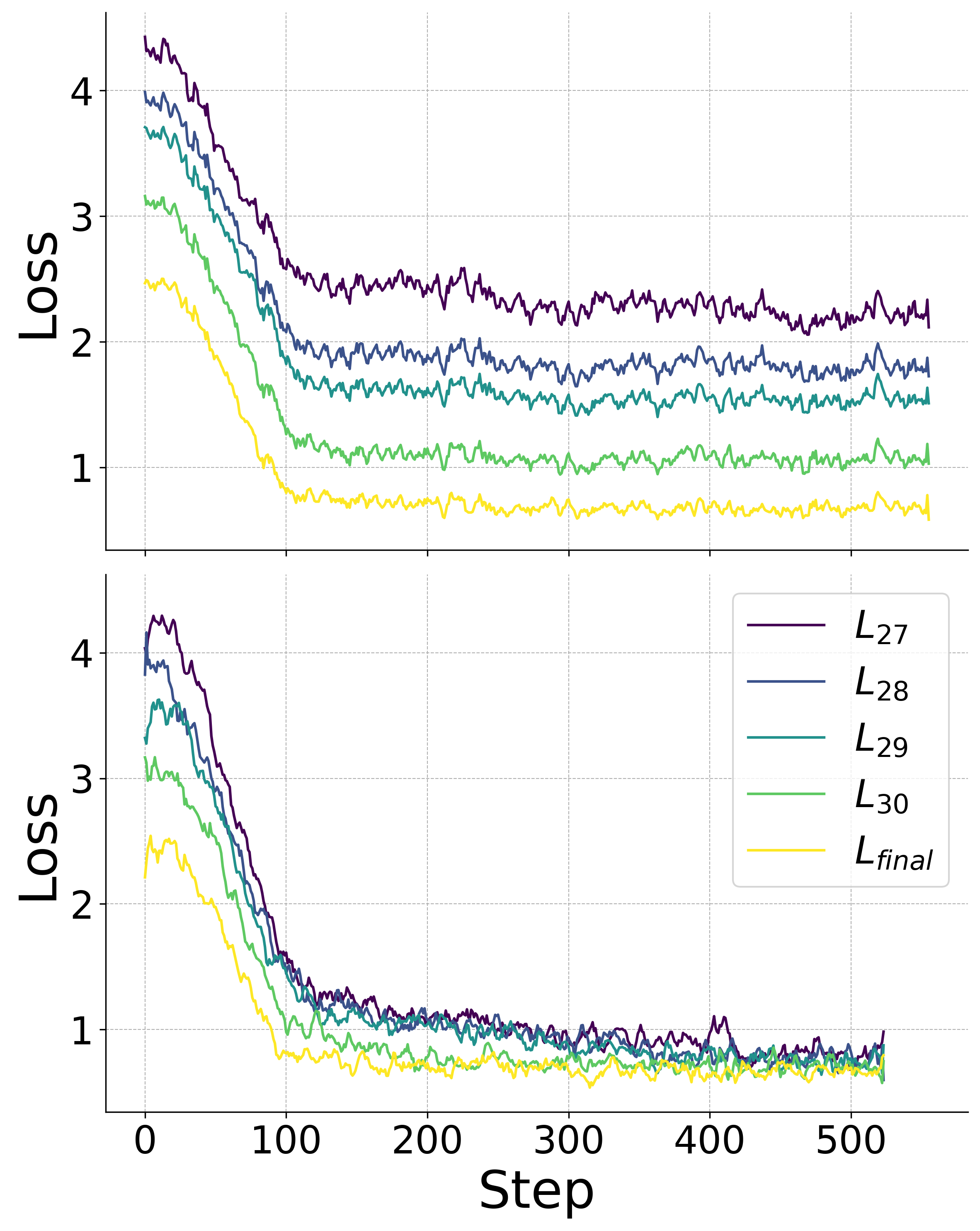}
        \caption{Tuning loss curves for LLaMA2-7B \citep{llama2} on ARC dataset \citep{arc}. Above shows the loss curve of late layers when optimising the loss based on the last layer output; Below shows the loss curves when optimising directly on each late layer output with the distillation signal from the last layer.\looseness-1}
        \label{fig:intro_loss}
        \vspace{-8mm}
    \end{wrapfigure}

\begin{figure*}[t]
    \centering
    \includegraphics[width=\linewidth]{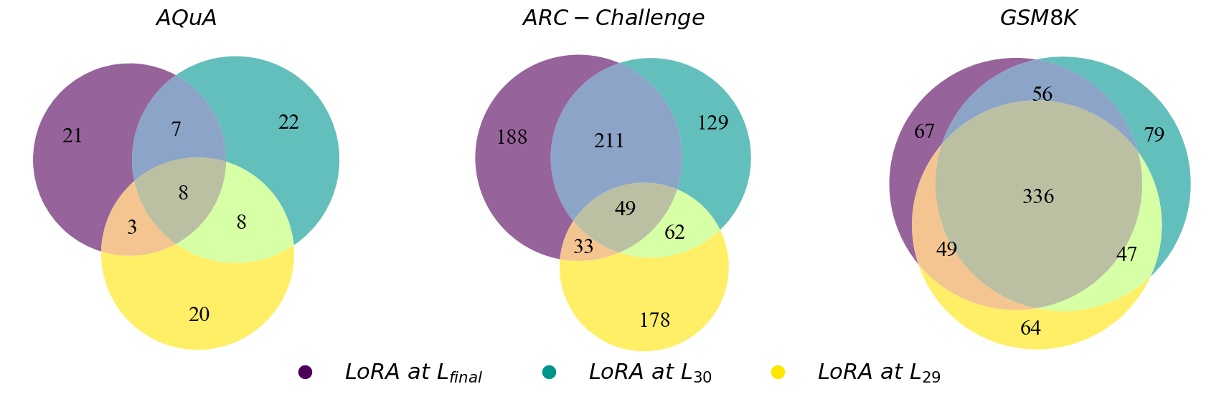}
    \vspace{-6mm}
    \caption{Intersection of solved problems by tuning late layers on the AQuA \citep{aqua}, ARC-Challenge \citep{arc}, and GSM8K \citep{gsm8k} datasets. The digits in the Venn diagram illustrate the number of overlapping solved problems and the complementary solved problems for each method.}
    \label{fig:intro_eval}
\end{figure*}

    In this work, we investigate the predictive power of the late layers\footnote{``Late'' layers often refer to those closer to the output, e.g., layers 25-32 in a 32-layer LLaMA 7B model in different literatures \citep{jumpingToConclusion, geva2023dissecting, meng2023locating}.}, which have proven to be task-aware in early exiting language models \citep{schuster2022confident, jumpingToConclusion}. 
    We begin by finetuning models directly on individual late layers by applying the pretrained language model heads to each layer's output to calculate the loss. Our initial observations at Figure~\ref{fig:intro_loss} indicate that the losses started at higher values, but eventually converged to similar levels with a simple distillation signal from the final layer. 
    Figure~\ref{fig:intro_eval} demonstrates that the trained ``models'' at these layers can even provide complementary evaluation results. These findings suggest that the late layers possess significant predictive potential. 
    
    Following this observation, we introduce the \textit{Mixture-of-Depths Ensemble} (MoDE) framework (\S \ref{sec:mod}).
    Unlike the ``mixture-of-experts'' paradigm, which uses different trained models as experts for processing different input tokens \citep{jiang2024mixtral},
    we propose a ``mixture'' across layers within a single model, where each layer output can be treated as a single model output. This approach allows us to add diversity and additional predictive power without significantly increasing parameters by training a simple gating network for the $i$-th late layer (\S \ref{sec:gate}).

    Experiments (\S \ref{sec:exp}) show that applying MoDE tuning yields modest but consistent gains on reasoning tasks with a minimal increase in trainable parameters (+0.04\%). Furthermore, when used in place of the late-layer LoRA modules, MoDE attains comparable performance while adding two orders of magnitude fewer trainable parameters ($+0.04\%$ versus $+10.3\%$). We frame these results as evidence that late-layer logits can be made directly useful for tuning at small overhead.
    To understand why MoDE helps and what it costs, we analyse its routing behaviour (\S \ref{sec:analysis}). The router relies mostly on late layers, but the per-layer normalisation and distillation signal also let earlier layers contribute. Wider ranges of ensemble help easy datasets yet hurt harder ones such as mathematical reasoning. Because the routing is sparse, activating only the top-scoring layers retains most of the accuracy while enabling early exit, speeding up generation by up to $1.6\times$.

    In summary, our contributions are:
    \begin{itemize}
        \item \textbf{Late-layer logits are a trainable predictive signal.} Optimised individually, the logits read off late layers reach losses comparable to the final layer and solve a complementary set of problems.
        \item \textbf{We introduce MoDE, a simple plug-in tuning mechanism with small overhead.} We ensemble late-layer logits through a learned router that composes with any existing tuning method and adds only negligible trainable parameters. MoDE matches much larger trainable modules at comparable cost, and, through sparse routing, can potentially speed up generation.
    \end{itemize}

\section{Mixture-of-Depths Ensemble}
\label{sec:mod}
Recent language models consist of an embedding layer, $n$ stacked transformer layers $L$, and an affine layer $\phi(\cdot)$ for predicting the next-word distribution, often referred to as the language model head \citep{geva2022transformer, luo2024understanding}. We aim to identify a layer range $k$, where the last $k$ layers carry higher-level task-aware information and can map the representations to meaningful predictive logits \citep{belrose2023eliciting}. 
For an LLM with $n$ layers, we define the set of the last $k$ layers as $\mathcal{K} = \{L_{n-k+1}, L_{n-k+2}, \dots, L_{n}\}$.
As shown in Figure \ref{fig:intro_loss}, late layers exhibit learning loss curves similar to the final layer, indicating their task informativeness. 
    
This range could also be set dynamically from inference-time signals; for instance, \citet{chuang2023dola} use the Jensen--Shannon divergence between early and final layers to pick more task-relevant layers for contrastive decoding. For simplicity, we instead fix a single $k$ across all tasks and leave dynamic selection to future work.

\subsection{Early-Exit for Late Layers}
\label{sec:early exit}
The idea of applying language heads directly to the hidden states of the middle layers, known as \textit{early exit} \citep{teerapittayanon2016branchynet, elbayad2020depth, schuster2022confident}, has proven effective even without a special training process \citep{kao2020bert}. The residual connections \citep{he2016deep} in transformer layers allow hidden representations to evolve gradually, enabling the formation of task-aware representations without abrupt changes.
    
Given a sequence of tokens $\{x_1, x_2, \dots, x_{t-1}\}$, the embedding layer first converts the tokens into a sequence of vectors $H_0=\{h_1^{(0)}, \dots, h_{t-1}^{(0)}\}$, where $h_{t}^{(0)} \in \mathbb{R}^{d}$ and $d$ is the hidden state dimension. This sequence $H_0$ is then processed successively by each transformer layer, with the output of the $j$-th layer denoted as $H_j$. The vocabulary head $\phi(\cdot)$ then outputs the logits $\ell_t$ of the next token $x_{t}$ over the vocabulary set $\mathcal{V}$:
\begin{align}
    \ell(x_{t} \mid x_{<t}) = \phi\bigl(\mathcal{N}_{p}(h_t^{(N)}))\bigr)_{x_{t}}, \quad x_{t} \in \mathcal{V}.
\end{align}
Here, $\mathcal{N}_{p}$ is the pre-trained normalisation module before the vocabulary head. This method is often considered a form of logit lens \citep{nostalgebraist2020logitlens}, which uses the vocabulary head to probe into inner representations.  While such projections are widely used for inference-time analysis and decoding, they are less commonly used as a trainable signal that feeds back into the final prediction. In \S \ref{sec:gate}, we show how to combine the train-time predictive power of late layers with the final-layer logits.

\begin{figure}[t]
    \centering
    \includegraphics[width=1\linewidth]{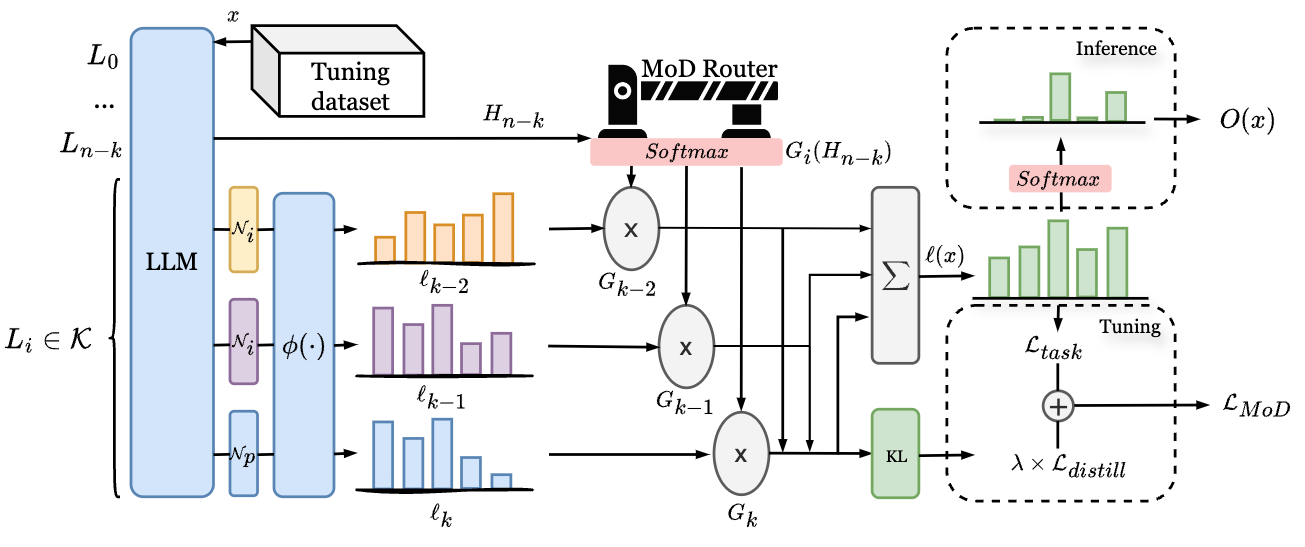}
    \caption{The overall framework of Mixture-of-Depths (MoDE), which can be applied on top of any tuning method like LoRA \citep{hu2021lora}. Given a pre-trained LLM and a tuning dataset, MoDE applies trainable normalisation $\mathcal{N}_k$ and pre-trained language model heads $\phi(\cdot)$ to the last $k$ layers $\{L_{n-k+1}, \ldots, L_n\}$. Each layer's output is combined using learned routing weights to produce the final logits. During training, a auxiliary teacher-enforced distillation loss $\mathcal{L}_{distill}$ is applied, where the final layer output serves as the teacher. MoDE utilises the ensemble logits during inference.}
    \vspace{-2mm}
    \label{fig:mod_main}
\end{figure}

\subsection{MoDE Routing Network}
\label{sec:gate}
Instead of applying $\phi(\cdot)$ only to the final layer, we route the most informative late-layer logits into the final prediction. Motivated by the MoE framework \citep{fedus2022review, jiang2024mixtral}, the ensemble logits are given by:
\begin{align}
    \sum_{i=0}^{k-1} G(x)_i \cdot \ell_i(x),~~~~G(x) := \text{Softmax}(x \cdot W_g).
\end{align}
Here, \( G(\cdot)_i \) is the routing weight for the $i$-th late layer, \( \ell_i(\cdot) \) is that layer's output logits, and the router input $x=H_{n-k}$ is the hidden state just before the last $k$ layers. The router is a single linear layer followed by a softmax.
The final logits are then obtained by summing the weighted logits from $k$ layers:
\begin{align}
    \ell(x_{t} \mid x_{<t}) = \sum_{i=0}^{k-1} G(x)_i \cdot \ell_i(x) 
\end{align}
Additionally, one advantage of the MoDE is its potential to improve inference efficiency by avoiding excessive computation when the routing vector is sparse. Following \citet{shazeer2017outrageously}, we achieve this by applying the softmax over the Top-K logits of the linear layer:
\begin{align}
    G_{\text{TopK}}(x) := \text{Softmax}(\text{TopK}(x \cdot W_g)),
\end{align}
where $(\text{TopK}(\ell))_i := \ell_i$ if $\ell_i$ is among the top-K coordinates of logits $\ell \in \mathbb{R}^k$ and $(\text{TopK}(\ell))_i := -\infty$ otherwise.

Our main experiments (\S \ref{sec:exp}) use the dense router \( G(x) \); we study the accuracy--efficiency trade-off of the sparse variant \( G_{\text{TopK}}(x) \) in \S \ref{sec:topk analysis}.

\subsection{Late Layers Adaptation by Normalisation and Distillation}
\label{sec:distill loss}
    
Directly combining the logits of late layers using the LM head can result in worse training loss at the start of tuning (Figure~\ref{fig:intro_loss}). Prior work \citep{belrose2023eliciting} learns a per-layer affine matrix $A_{\ell}$ mapping hidden states into the LM head's input space, but this adds many additional parameters. We seek a cheaper adaptation that interferes little with the model's predictions.
    
Inspired by normalisation studies in neural networks and the effectiveness of tuning the normalisation module for domain adaptation \citep{zhao2023tuning}, we propose tuning an additional normalisation module for each late layer as a simple yet powerful adaptation method. We set the additional normalisation module $\mathcal{N}_k$ to match the architecture of the pretrained $\mathcal{N}_p$. For instance, in the LLaMA family \citep{llama2,llama3}, we follow the LayerNorm setting \citep{ba2016layer}. The learnable parameters in the normalisation, $\gamma_{k}$ and $\beta_{k}$, are trained individually for each $k$-th late layer to ensure specific adaptation for each layer.
    
Following \S \ref{sec:gate}, we treat each of the $k-1$ earlier late layers as an initially weaker model and use the final layer as a teacher. A distillation loss pulls each earlier layer's output distribution $P_i$ towards the final distribution $P_n$, computed as the sum of their KL divergences:
\begin{align}
    \mathcal{L}_{distill} = \sum_{i=0}^{k-2} \text{KL}(P_i \parallel P_n),
\end{align}
where $P_i$ is the output distribution of layer $i$, and $P_n$ is the output distribution of the final layer. The final loss is then the sum of the task loss and the distillation loss:
\begin{align}
    \mathcal{L}_{\text{MoDE}} = \mathcal{L}_{task} + \lambda \mathcal{L}_{distill},
\end{align}
where $\lambda$ is a hyperparameter that controls the weight of the distillation loss.
By tuning with the normalisation modules and distillation loss, we adapt the $k-1$ layer representations to be more suitable for language modelling, ensuring their contributions are aligned with the original task loss.
    
\section{Experiments}
\label{sec:exp}


    
We evaluate MoDE on arithmetic reasoning, commonsense reasoning, and instruction following. Because MoDE only reads the late-layer logits and leaves hidden-state dimensions unchanged, it can be layered on top of any tuning method; we therefore report it both on its own and combined with existing parameter-efficient finetuning (PEFT) methods.

\paragraph{Baselines.} We denote a single LoRA layer as $L_{\text{LoRA}}$ and compare against four PEFT baselines chosen to span the two dominant PEFT families, so that we can test whether MoDE's late-layer signal adds value \emph{on top of} strong methods rather than only over a weak one:
\begin{itemize}
    \item $\text{LoRA}_{\neg\mathcal{K}}$: LoRA \citep{hu2021lora} applied to every transformer layer \emph{except} the last $|\mathcal{K}|$ late layers. This isolates the late layers' contribution and is the base to which the lightweight MoDE variant is added.
    \item $\text{LoRA}_{\text{ALL}}$: LoRA on all layers, identical to $\text{LoRA}_{\neg\mathcal{K}} + L_{\text{LoRA}}\times|\mathcal{K}|$. This is the natural baseline that simply adds trainable LoRA modules to the late layers.
    \item DoRA \citep{liu2024doraweightdecomposedlowrankadaptation}: a stronger \emph{weight-based} method that decomposes each weight into magnitude and direction to better match full finetuning.
    \item ReFT \citep{wu2024reft}: a \emph{representation-based} method that learns interventions directly on hidden states, and is conceptually the closest baseline to MoDE.
\end{itemize}
We select DoRA and ReFT because they are the strongest recent PEFT methods on this benchmark and represent the weight-editing and representation-editing families, respectively. MoDE is applied on top of $\text{LoRA}_{\text{ALL}}$ and DoRA; we also report the minimal $\text{LoRA}_{\neg\mathcal{K}}+\text{MoDE}_{\mathcal{K}}$ variant, which adds only $+0.04\%$ trainable parameters. Full baseline configurations and hyperparameters are given in Appendix~\ref{app:baselines}.

\paragraph{Setup.} We run all experiments on LLaMA2-7B \citep{llama2} and LLaMA3-8B \citep{llama3}. The distillation weight $\lambda$ is set to $0.0001$ for all datasets and models, and the router is Gaussian-initialised (mean $0$, standard deviation $0.02$). All runs use NVIDIA A6000 GPUs; further settings are in Appendix~\ref{app:setting}.

\subsection{Arithmetic Reasoning}
\label{exp:math}
\begin{table}[t]
\caption{
Accuracy comparison of MoDE with baselines on seven arithmetic reasoning datasets.
$\text{LoRA}_{\neg\mathcal{K}}$ denotes LoRA applied to all transformer layers except the last $|\mathcal{K}|$ late layers, and
$\text{LoRA}_{\text{ALL}}$ denotes LoRA applied to all layers.
DoRA and ReFT are standard PEFT baselines.
Bold and underlined values indicate the best and second-best result within each model block, respectively.
}
\label{tab:arithmetic_reasoning}
\begin{center}
\begin{sc}
\resizebox{\textwidth}{!}{
\begin{tabular}{lcccccccc}
\toprule
\textbf{Method} & \textbf{AddSub} & \textbf{AQuA} & \textbf{GSM8K} & \textbf{MAWPS} & \textbf{MultiArith} & \textbf{SingleEq} & \textbf{SVAMP} & \textbf{Avg.} \\
\midrule

\textit{LLaMA2-7B} & & & & & & & & \\
\cmidrule(l){1-1}

$\text{LoRA}_{\neg\mathcal{K}}$
& 49.1 & 24.0 & 42.4 & 61.7 & 81.4 & 66.1 & 46.6 & 53.0 \\

$\text{LoRA}_{\text{ALL}}$~\citep{hu2021lora}
& 51.1 & 24.4 & 43.6 & 62.6 & 84.2 & 66.9 & 47.7 & 54.5 \\

$\text{DoRA}$~\citep{liu2024doraweightdecomposedlowrankadaptation}
& \underline{51.4} & 24.8 & \underline{44.0} & 62.9 & \underline{84.5} & \underline{67.4} & \underline{48.2} & 54.7 \\

$\text{ReFT}$~\citep{wu2024reft}
& 49.0 & \textbf{25.8} & 36.5 & 59.6 & 80.5 & 63.8 & 44.0 & 51.3 \\

$\text{LoRA}_{\text{ALL}}+\textbf{MoDE}_{\mathcal{K}}$
& 51.2 & \underline{25.5} & 43.9 & 63.1 & 84.3 & 67.3 & 48.0 & \underline{54.8} \\

$\text{DoRA}+\textbf{MoDE}_{\mathcal{K}}$
& \textbf{51.6} & \textbf{25.8} & \textbf{44.4} & \underline{63.4} & \textbf{84.6} & \textbf{67.6} & \textbf{48.4} & \textbf{55.1} \\

\rowcolor{gray!8}
$\text{LoRA}_{\neg\mathcal{K}}+\textbf{MoDE}_{\mathcal{K}}$ (+0.04\%)
& 50.1 & 24.3 & 43.4 & \textbf{63.7} & 82.2 & 66.8 & 47.5 & 54.0 \\

\midrule

\textit{LLaMA3-8B} & & & & & & & & \\
\cmidrule(l){1-1}

$\text{LoRA}_{\neg\mathcal{K}}$
& 91.8 & 25.0 & 60.6 & 90.2 & 98.8 & 95.7 & 72.8 & 76.4 \\

$\text{LoRA}_{\text{ALL}}$~\citep{hu2021lora}
& \underline{92.7} & 25.6 & 61.6 & 90.8 & \textbf{99.5} & 96.3 & 73.8 & 77.2 \\

$\text{DoRA}$~\citep{liu2024doraweightdecomposedlowrankadaptation}
& 92.2 & 26.8 & \underline{62.7} & 91.2 & 98.8 & \underline{96.9} & 74.0 & 77.5 \\

$\text{ReFT}$~\citep{wu2024reft}
& 91.0 & \textbf{27.5} & 54.0 & 88.5 & 97.0 & 94.0 & 67.5 & 74.2 \\

$\text{LoRA}_{\text{ALL}}+\textbf{MoDE}_{\mathcal{K}}$
& \textbf{93.0} & 26.2 & 62.2 & \underline{91.3} & \textbf{99.5} & 96.5 & \underline{74.2} & \underline{77.6} \\

$\text{DoRA}+\textbf{MoDE}_{\mathcal{K}}$
& 92.4 & \underline{27.4} & \textbf{63.1} & \textbf{91.6} & \underline{99.0} & \textbf{97.1} & \textbf{74.7} & \textbf{77.9} \\

\rowcolor{gray!8}
$\text{LoRA}_{\neg\mathcal{K}}+\textbf{MoDE}_{\mathcal{K}}$ (+0.04\%)
& 92.2 & 25.5 & 61.5 & 90.8 & 98.8 & 96.0 & 73.5 & 76.9 \\

\bottomrule
\end{tabular}
}
\end{sc}
\end{center}
\end{table}
    Arithmetic reasoning includes seven datasets for math word problems: AddSub \citep{addsub}, AQuA \citep{aqua}, GSM8K \citep{cobbe2021training}, MAWPS \citep{koncel-kedziorski-etal-2016-mawps}, MultiArith \citep{mutli_arith}, SingleEq \citep{singleeq}, and SVAMP \citep{svamp}. Models need to generate chain-of-thought \citep{wei2022chain} reasoning steps before the final answer. We replicate the experimental setup from \citet{hu2023llmadapters} on a combined dataset of these seven arithmetic reasoning tasks with LM-generated chain-of-thought steps (\textsc{Math7K}) and report scores on all test sets. We only evaluate the correctness of the final numeric or multiple-choice answer. Details of the dataset are provided in Appendix \ref{app:dataset math}.
    For \textsc{Math7K}, we set $k$ to 3 for both LLaMA2-7B and LLaMA3-8B across all datasets. Different models and datasets may benefit from a different $k$, or from selecting $k$ dynamically during training, which we leave to future work.

    Table~\ref{tab:arithmetic_reasoning} shows that MoDE consistently improves whichever base method it augments. On both models, adding $\text{MoDE}_{\mathcal{K}}$ to $\text{LoRA}_{\text{ALL}}$ and to DoRA raises their averages, and $\text{DoRA}+\text{MoDE}_{\mathcal{K}}$ is the best configuration overall (55.1 on LLaMA2-7B, 77.9 on LLaMA3-8B). The lightweight $\text{LoRA}_{\neg\mathcal{K}}+\text{MoDE}_{\mathcal{K}}$ variant, which adds only $+0.04\%$ parameters, still beats its $\text{LoRA}_{\neg\mathcal{K}}$ base and comes within $0.3$--$0.5$ points of the far larger $\text{LoRA}_{\text{ALL}}$, indicating that the late-layer logits alone carry most of the signal behind these gains. ReFT trails on arithmetic, suggesting that representation interventions are less suited to multi-step numeric reasoning than logit-level ensembling.
    
\subsection{Commonsense Reasoning and General Language Modelling}
\label{exp:commonsense}
    Commonsense reasoning is evaluated using four datasets: the Challenge Set and Easy Set of ARC \citep{arc}, BoolQ \citep{boolq}, and OBQA \citep{OpenBookQA2018}. These tasks are formulated as multiple-choice problems. We follow the setup from \citet{hu2023llmadapters}, but train each dataset separately to assess the effectiveness of our MoDE framework on individual datasets. 
    To evaluate general language modelling capability, we select 20\% of the TruthfulQA dataset and report the True*Informative score. We also report the performance on the STEM subtasks of the MMLU benchmark, following the setup of \citet{GPT3}. 
    \begin{wraptable}{r}{0.48\textwidth}
\centering
\vspace{-4mm}
\caption{
Average MT-Bench scores assigned by GPT-4o to answers generated by tuned models.
The +Params column reports trainable parameters as a percentage of the full backbone parameter count.
Bold values indicate the best result within each model block.
}
\vspace{+1mm}
\resizebox{\linewidth}{!}{
\begin{tabular}{llcc}
\toprule
\textbf{Model} & \textbf{Method} & \textbf{+Params (\%)} & \textbf{Score} \\
\midrule

\multirow{4}{*}{LLaMA2-7B}
& $\text{LoRA}$
& 2.37 & 5.7 \\

& $\text{DoRA}$
& 2.39 & \underline{6.0} \\

& $\text{LoRA}+\textbf{MoDE}_{\mathcal{K}}$
& 2.37 & 5.8 \\

& $\text{DoRA}+\textbf{MoDE}_{\mathcal{K}}$
& 2.39 & \textbf{6.1} \\

\midrule

\multirow{4}{*}{LLaMA3-8B}
& $\text{LoRA}$
& 2.09 & 7.4 \\

& $\text{DoRA}$
& 2.11 & \underline{7.5} \\

& $\text{LoRA}+\textbf{MoDE}_{\mathcal{K}}$
& 2.09 & \underline{7.5} \\

& $\text{DoRA}+\textbf{MoDE}_{\mathcal{K}}$
& 2.11 & \textbf{7.6} \\

\bottomrule
\end{tabular}
}
\label{tab:mtbench}
\end{wraptable}
    Dataset details are provided in Appendix \ref{app:dataset commonsense}; we keep the settings of \S \ref{exp:math}. As shown in Table~\ref{tab:commonsense_reasoning} (full results in Appendix~\ref{app:commonsense_results}), MoDE again improves every base method it augments: $\text{LoRA}_{\text{ALL}}+\text{MoDE}_{\mathcal{K}}$ and $\text{DoRA}+\text{MoDE}_{\mathcal{K}}$ beat $\text{LoRA}_{\text{ALL}}$ and DoRA on both models, and the $+0.04\%$ variant improves over $\text{LoRA}_{\neg\mathcal{K}}$ by up to $3.2$ points (62.4$\rightarrow$65.6 on LLaMA2-7B). Here ReFT is the strongest standalone method; we attribute this to its heavier representation editing being well matched to these multiple-choice tasks. MoDE does not overtake ReFT, but it delivers a reliable additive gain on top of the weight-based methods at a fraction of their parameters. Taken together, the pattern across both task types supports our claim that late-layer logits are a complementary, broadly applicable training signal.

\subsection{Instruction Following}
\label{inst follow exp}
    We finetune LLaMA2-7B and LLaMA3-8B on a 10K subset of the cleaned Alpaca dataset \citep{taori2023alpaca} and evaluate on MT-Bench \citep{zheng2023judging}, where GPT-4o scores responses to 80 multi-turn questions out of 10.

    Table~\ref{tab:mtbench} shows that adding MoDE gives small but consistent improvements over both LoRA and DoRA at the same $+0.04\%$ overhead (e.g.\ $6.0\rightarrow6.1$ on LLaMA2-7B and $7.5\rightarrow7.6$ on LLaMA3-8B for DoRA). The gains are smaller than on reasoning tasks, consistent with instruction following depending more on late-layer formatting that MoDE's ensembling only partly captures. We view stronger instruction-tuning adaptations of MoDE as future work.\looseness-1

\section{Analysis}
\label{sec:analysis}
    Using the training setup from \S \ref{sec:exp}, we analyse \emph{where} MoDE's gains come from and \emph{what} they cost. Three findings emerge. (i) The router concentrates weight on the last one or two layers, but the trainable normalisation and distillation modules let earlier late layers contribute, shifting the learned routing pattern and improving over the baseline either way (\S \ref{sec:sparsity}). (ii) The useful ensemble size is small: $k\!\approx\!3$--$4$ is best overall, larger ranges help easy tasks (ARC-e) but hurt harder ones (GSM8K), so more late layers are not always better (\S \ref{sec:topk analysis}). (iii) Sparse Top-$K$ routing recovers most of the accuracy while enabling early exit, giving up to $1.6\times$ faster generation and exposing a tunable accuracy--efficiency trade-off (\S \ref{sec:topk analysis}). An ablation confirms that both adaptation components are beneficial (\S \ref{sec:ablation}).

\subsection{Learned Routing Pattern Across Tokens}
\label{sec:sparsity}
    
    
    We study how the router allocates weight across the $\mathcal{K}$ ensemble layers during training. For each route we measure \emph{sparsity}, the fraction of routing weights below $\epsilon=1\times10^{-5}$; lower sparsity means the route is used more selectively (per-route mean and variance are reported in Appendix~\ref{app:analysis1}). We compare MoDE trained on top of LoRA against MoDE trained without the $k$ late-layer LoRA modules, using LLaMA2-7B on ARC-e.
    \begin{figure}[t]
        \centering
        \includegraphics[width=0.9\textwidth]{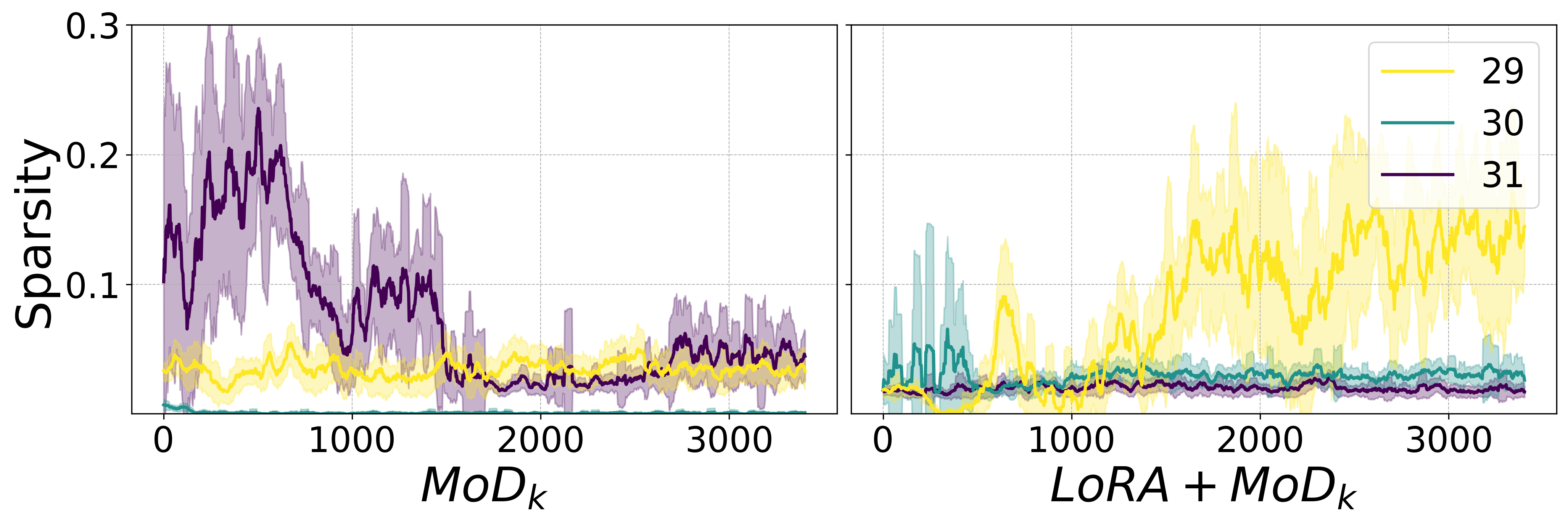} 
        \caption{Sparsity scores for MoDE (left) and MoDE trained with $k$ LoRA layers (right). The curve is smoothed using moving average smoothing.}
        \label{fig:sparsity}
    \end{figure}
    Figure~\ref{fig:sparsity} shows that the two settings learn different routing patterns. Without the $k$ LoRA layers, the router relies mainly on the last two layers---the final layer stays least sparse throughout training. With the $k$ LoRA layers, the final layer becomes much sparser while the earlier late layers stay active, indicating that MoDE's trainable modules let earlier layers contribute more to the ensemble (\S \ref{sec:distill loss}). Both settings outperform the baseline (Table~\ref{tab:commonsense_reasoning}), so different weight combinations can each add predictive signal.
 
\subsection{MoDE Sparse Routing}
\label{sec:topk analysis}
    
The high sparsity observed in \S \ref{sec:sparsity} suggests the router can be made sparse at inference. We therefore test the $G_{\text{TopK}}$ variant (\S \ref{sec:gate}): if routing stays sparse without hurting the ensemble, selecting only the top routes before the last layer enables early exit and faster generation.
\begin{figure}[t]
    \centering
    \includegraphics[width=\linewidth]{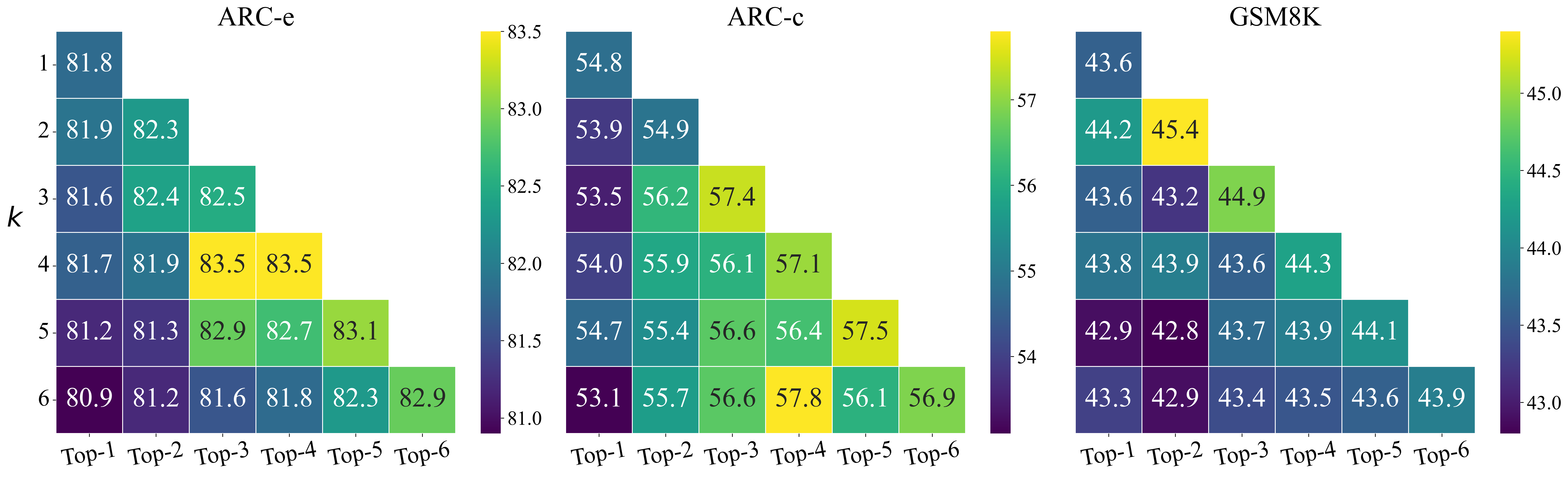} 
    \caption{Accuracy scores for different $k$ ensemble layer ranges and Top-K sparse routing values. Lighter colors indicate better performance.}
    \label{fig:top6}
\end{figure}
We call this $\text{MoDE}_{sparse}$ and use a larger range $k=6$ to maximise early-exit opportunities; results for other datasets are in Appendix \ref{app:topk}.
\begin{wraptable}{r}{0.5\textwidth}
\vspace{-5mm}
\caption{Acceleration ratios for different Top-K values when $k=6$ compared to the LoRA baseline. The results represent the overall speedup across 1000 iterations for each dataset.}
\centering
\resizebox{0.9\linewidth}{!}{
\begin{tabular}{c|cccc}
    \toprule
     Dataset & Top-2 & Top-3 & Top-4 & Top-5 \\
    \midrule
    ARC-e & $1.4\times$ & \bf $1.5\times$ & $1.4\times$ & $1.1\times$ \\
    ARC-c & \bf $1.6\times$ & $1.4\times$ & $1.3\times$ & $1.0\times$\\
    \bottomrule
\end{tabular}
}
\label{tab:accelerate}
\vspace{-1mm}
\end{wraptable}
We ask two questions. First, does a larger ensemble range $k$ help or add noise? Figure \ref{fig:top6} shows the optimum is around $k=3$--$4$: increasing $k$ steadily helps easy datasets (ARC-e) but hurts reasoning-heavy ones (GSM8K at $k=6$). Second, does Top-$K$ activation hurt the ensemble? The dense router ($k=\text{Top-}K$) is always best, and although Top-$K$ slightly lowers accuracy it still beats the $k=1$ baseline. Meanwhile, Table \ref{tab:accelerate} shows that larger Top-$K$ yields greater speedups, revealing a tunable trade-off between MoDE's predictive power and its efficiency.

\subsection{Ablation Study on Adaptation Modules}
\label{sec:ablation}
\begin{wrapfigure}{r}{0.5\textwidth}
    \centering
    \includegraphics[width=0.48\textwidth]{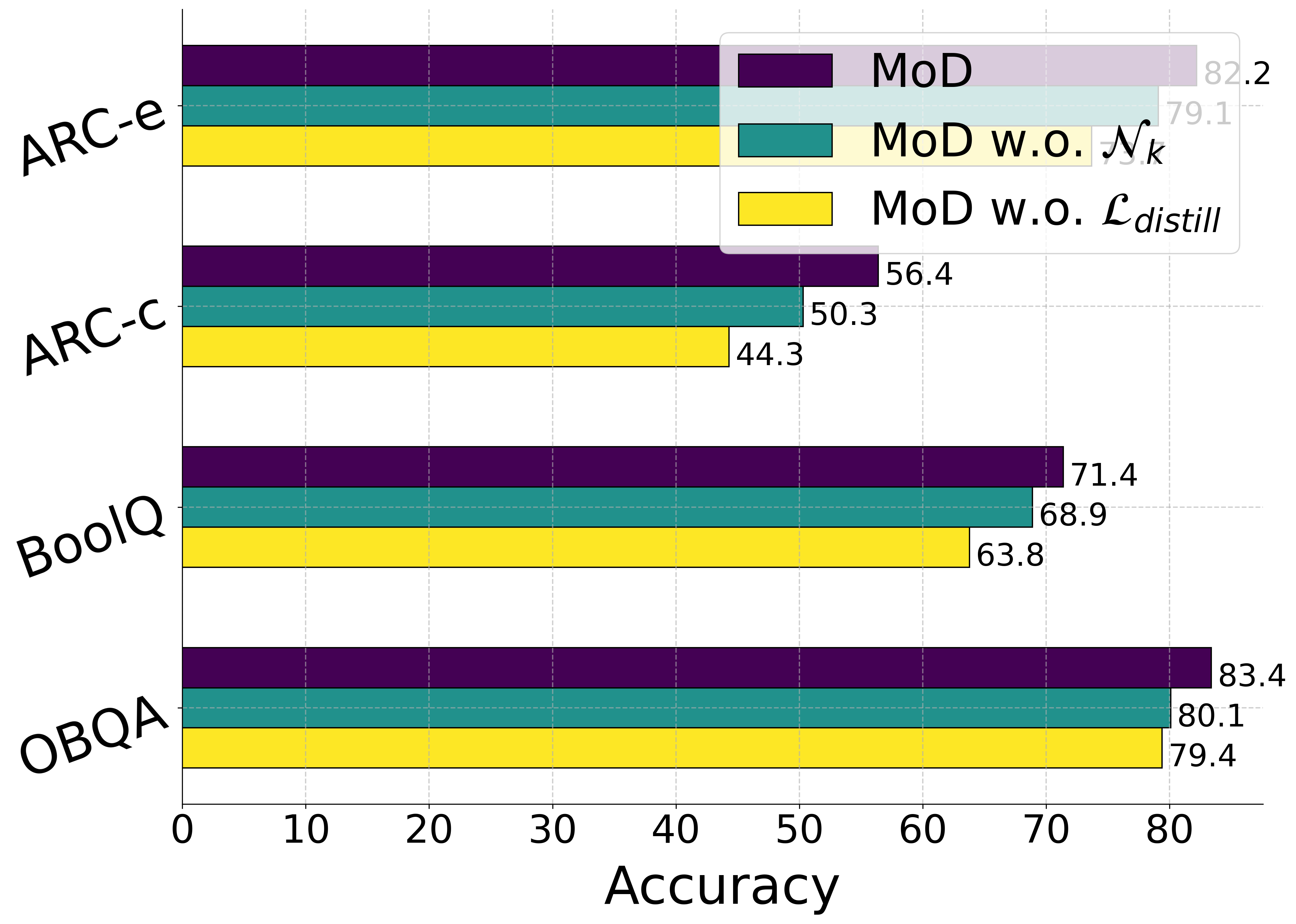}
    \caption{Ablation study results for MoDE on four commonsense reasoning datasets using the LLaMA2-7B model.}
    \label{fig:ablation}
\end{wrapfigure}
Figure \ref{fig:intro_loss} shows that adding $\mathcal{L}_{distill}$ stabilises the late-layer logits within the first 200 steps of training, even though the pretrained model was never trained to read predictions off these layers. We ablate the two adaptation components from \S \ref{sec:distill loss}:
1) \textbf{MoDE w/o $\mathcal{N}_k$}: replace the per-layer trainable normalisation with the pretrained head normalisation for all $k$ ensemble layers.
2) \textbf{MoDE w/o $\mathcal{L}_{distill}$}: drop the distillation loss and train on the task (cross-entropy) loss alone.
On four commonsense datasets with LLaMA2-7B (Figure \ref{fig:ablation}), removing either component hurts performance, and the distillation loss matters more: its last-layer supervision appears essential for the ensemble layers to adapt. Other supervision signals, such as JS divergence \citep{chuang2023dola} or reinforcement learning \citep{wu2024mixtureofskills}, are left for future work.
    
\section{Related Work}

\noindent \textbf{Reading and Training Intermediate Layers} A line of interpretability work reads predictions directly out of intermediate layers: the logit lens \citep{nostalgebraist2020logitlens} projects hidden states through the LM head, and the tuned lens \citep{belrose2023eliciting} adds per-layer affine corrections. These probes show that late-layer representations already encode task-relevant, near-final predictions \citep{li2024inference, geva2023dissecting}, which downstream methods exploit at inference time to improve factuality \citep{chuang2023dola} or efficiency through early exit \citep{xin-etal-2020-deebert, schuster2022confident}. A few methods also \emph{train} intermediate layers---LayerSkip \citep{elhoushi2024layerskip} adds an early-exit loss and Mixture-of-Depths \citep{raposo2024mixture} routes tokens to skip layers---but these target inference efficiency. Parameter-efficient finetuning offers a complementary way to adapt LLMs cheaply, either by editing weights, as in LoRA \citep{hu2021lora} and DoRA \citep{liu2024doraweightdecomposedlowrankadaptation}, or by editing hidden representations, as in ReFT \citep{wu2024reft}; we use these three as baselines. In contrast to all of the above, MoDE makes the late-layer \emph{logits} actionable for tuning: it feeds their ensemble back into the training objective rather than only probing, decoding, or editing weights, and it composes on top of any of these finetuning methods.

\noindent \textbf{Logit-Level Arithmetic} Operating at the logit level is effective for steering LLM outputs \citep{luo2024understanding}. Across models, ensembling or contrasting logits from different LMs improves generation in line with the Mixture-of-Experts paradigm \citep{shazeer2017outrageously, jiang2024mixtral, liu-etal-2021-dexperts, gera-etal-2023-benefits}; within a single model, contrasting logits across layers improves trustworthiness and mitigates the cost of larger models \citep{chuang2023dola, liu-etal-2024-tuning}. We likewise ensemble logits, but uniquely turn a single model's late layers into a \emph{training} signal, whereas this prior work applies logit ensembling only at inference.\looseness-1

\section{Discussion and Conclusion}
\label{sec:conclusion}
\textbf{Limitations.} MoDE relies on a fixed late-layer range $k$ chosen empirically; dynamic, per-task selection of $k$ and better tuning of the distillation weight $\lambda$ remain open. Its gains are smaller for instruction following than for reasoning (\S \ref{inst follow exp}), and our experiments are limited to decoder-only LLMs; extending MoDE to broader model families and to larger models is left for future work.

\textbf{Conclusion.} Starting from the interpretability observation that late-layer logits already carry task-relevant, near-final predictions, we showed this signal can be made actionable for tuning and introduced \textit{Mixture-of-Depths Ensemble} (MoDE), which ensembles late-layer logits through a learned router with lightweight adaptation. MoDE composes with any tuning method and yields modest but consistent reasoning gains at negligible overhead, and can stand in for substantially larger trainable modules at comparable performance. We present it as a mechanism study---a lightweight, complementary direction for turning intermediate-layer representations into a useful training signal.

\bibliography{colm2026_conference}
\bibliographystyle{colm2026_conference}

\appendix
\section{Datasets}
\label{app:dataset}
    \begin{table}[h]
\centering
\caption{Details of 11 datasets being evaluated according to \citet{hu2023llmadapters} and \citet{hendrycks2021measuringmassivemultitasklanguage}. Math: arithmetic reasoning. CS: commonsense reasoning.}
\vspace{2mm}
\begin{sc}
\begin{tabular}
{lcccr}
\toprule
Dataset & Domain & \# train & \# test &Answer  \\ \midrule
{MultiArith} & Math   & -    &600      &Number \\
{AddSub}     & Math   & -    &395      &Number \\
{GSM8K}      & Math   & 8.8K &1,319    &Number \\
{AQuA}       & Math   & 100K &254      &Option \\
{SingleEq}   & Math   & -    &508      &Number \\
{SVAMP}      & Math   & -    &1,000    &Number \\
{MAWPS}      & Math   & -    &238    &Number \\
{BoolQ}      & CS     & 9.4K &3,270    &Yes/No \\
{ARC-e}      & CS     & 2.3K &2,376    &Option \\
{ARC-c}      & CS     & 1.1K &1,172    &Option \\
{OBQA}       & CS     & 5.0K &500      &Option \\
{HellaSwag}  & CS     & 39.9K &10042      &Option \\
{MMLU}       & -     & 99.8K &14042      &Option \\
\bottomrule
\end{tabular}
\label{tab:dataset_description}
\end{sc}
\end{table}
    \noindent \textbf{Dataset Statistics and Examples} Dataset statistics and simplified training examples from each dataset are provided in Table \ref{tab:dataset_description}. The original training dataset of Math10K accidentally includes testing examples from AddSub, MultiArith, and SingleEq tasks, as these tasks are part of the MAWPS training dataset, causing a data leak. To address this, we replicate the experimental setup suggested by \citet{hu2023llmadapters} on a combined training dataset (\textsc{Math7K}). For the commonsense reasoning dataset, we trained individual datasets with a newly designed prompt format to address various issues reported with the LLaMA tokenizer in the original prompt format.

\subsection{Arithmetic Reasoning}
\label{app:dataset math}
    We conduct extensive empirical studies on fourteen benchmark datasets, focusing on two categories of reasoning problems:
    \textbf{Arithmetic Reasoning:} 
    1. \textbf{GSM8K} \citep{gsm8k}: A dataset comprising high-quality, linguistically diverse grade school math word problems created by human problem writers.
    2. \textbf{SVAMP} \citep{svamp}: A benchmark of one-unknown arithmetic word problems designed for up-to-4th grade students, created by making simple modifications to problems from an existing dataset.
    3. \textbf{MultiArith} \citep{mutli_arith}: A dataset featuring math word problems that require multiple reasoning steps and operations.
    4. \textbf{AddSub} \citep{addsub}: A collection of arithmetic word problems focused on addition and subtraction.
    5. \textbf{AQuA} \citep{aqua}: A dataset of algebraic word problems accompanied by natural language rationales.
    6. \textbf{SingleEq} \citep{singleeq}: A set of grade-school algebra word problems that map to single equations of varying lengths.
    7. \textbf{MAWPS} \citep{koncel-kedziorski-etal-2016-mawps}: A collection of math word problems assembled from several earlier datasets to span a range of difficulty levels.

    \begin{figure*}[t]
        \centering
        \includegraphics[width=0.9\textwidth]{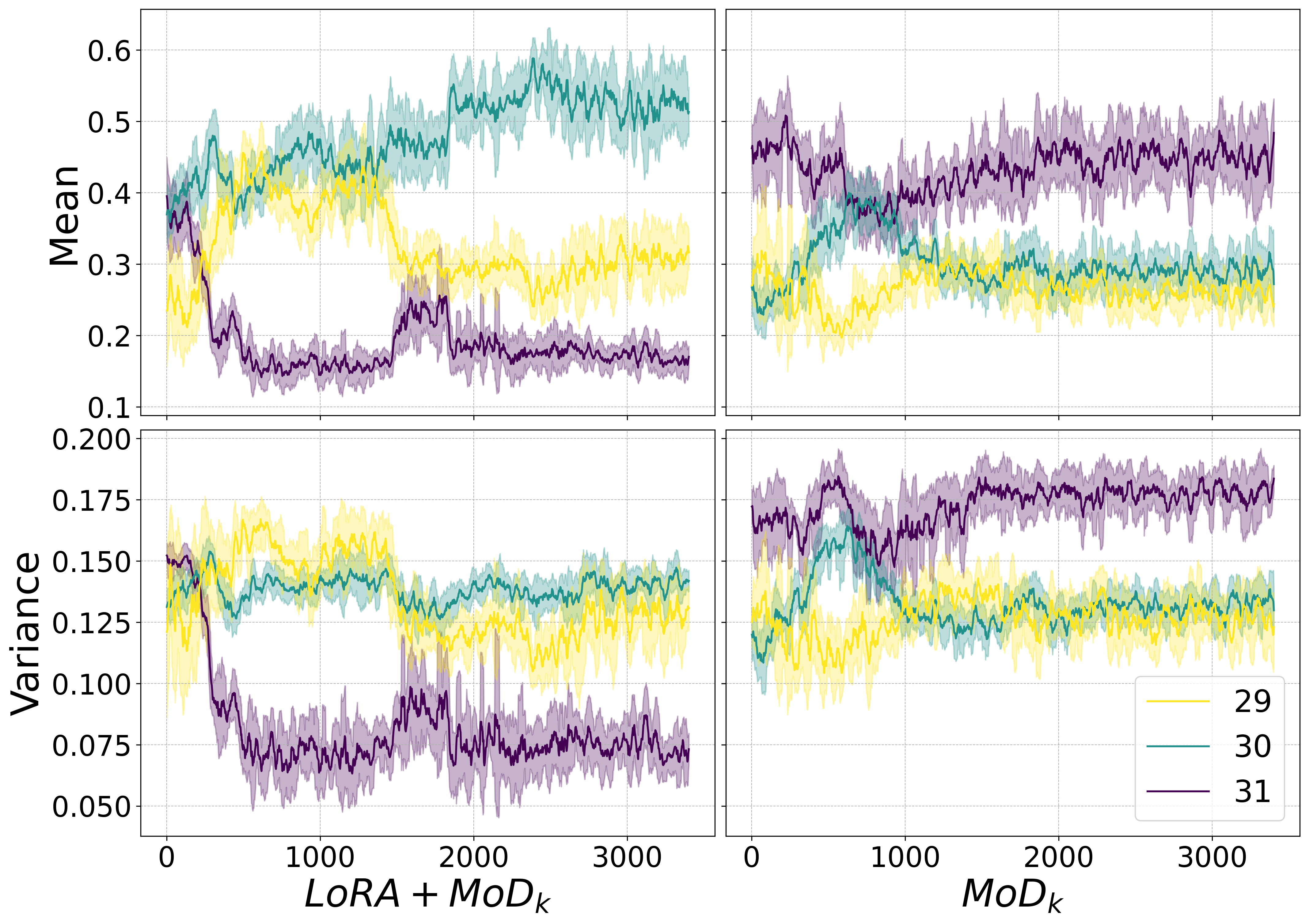} 
        \caption{Mean and variance for MoDE (right) and MoDE trained with $k$ LoRA layers (left). The curve is smoothed using moving average smoothing with a window size of 3 and $k=3$.}
        \label{fig:meanvar}
    \end{figure*}

\subsection{Commonsense Reasoning}
\label{app:dataset commonsense}
    We trained our method on four commonsense reasoning dataset separately. They are:
    1. \textbf{BoolQ} \citep{boolq}: A question-answering dataset containing 15,942 naturally occurring yes/no questions generated in unprompted and unconstrained settings.
    2. \textbf{ARC-c} and \textbf{ARC-e} \citep{arc}: The Challenge Set and Easy Set of the ARC dataset, consisting of genuine grade-school level, multiple-choice science questions.
    3. \textbf{OBQA} \citep{OpenBookQA2018}: A dataset containing questions that require multi-step reasoning, use of additional common and commonsense knowledge, and rich text comprehension.

\section{Experiment Settings}
\label{app:setting}
    We mainly follow the experimental settings of \citet{hu2023llmadapters}. We maintain a batch size of 16 and set the learning rate for all methods to 3e-4. Each method is fine-tuned for two epochs on each dataset.

\section{Baselines}
\label{app:baselines}
    We compare MoDE against four parameter-efficient finetuning (PEFT) baselines. They span the two dominant PEFT families---weight-based and representation-based adaptation---so that we can test whether MoDE adds value on top of strong methods, not only over a weak one. Unless stated otherwise, all baselines share the optimisation settings in Appendix~\ref{app:setting} (batch size 16, learning rate 3e-4, two epochs).

    \noindent\textbf{LoRA} \citep{hu2021lora} freezes the pretrained weights and learns a low-rank update $\Delta W = BA$ for selected projection matrices. We use it as the base tuning method and define two configurations: $\text{LoRA}_{\text{ALL}}$ applies LoRA to every transformer layer, while $\text{LoRA}_{\neg\mathcal{K}}$ excludes the last $|\mathcal{K}|$ late layers so that the contribution of the late layers can be isolated. We use rank $r=32$ and $\alpha=64$ on the query, key, value, and output projections, following \citet{hu2023llmadapters}.

    \noindent\textbf{DoRA} \citep{liu2024doraweightdecomposedlowrankadaptation} decomposes each pretrained weight into a magnitude vector and a directional matrix, and applies a LoRA-style low-rank update to the direction. This narrows the gap to full finetuning while keeping the parameter budget close to LoRA. We use the same rank and target modules as LoRA. We select DoRA as a representative \emph{state-of-the-art weight-based} baseline and also as a base method that MoDE is layered on top of.

    \noindent\textbf{ReFT} \citep{wu2024reft} keeps the model weights frozen and instead learns low-rank interventions on hidden representations at selected layers and positions (we use the LoReFT variant). It is the strongest \emph{representation-based} PEFT method on this benchmark and is conceptually the closest baseline to MoDE, since both operate on intermediate representations rather than weights; the key difference is that ReFT edits hidden states whereas MoDE ensembles the resulting late-layer logits into the final prediction. We use the authors' recommended intervention rank and layer configuration.

    \noindent\textbf{MoDE configurations.} MoDE is applied on top of $\text{LoRA}_{\text{ALL}}$ and DoRA, denoted $\text{LoRA}_{\text{ALL}}+\text{MoDE}_{\mathcal{K}}$ and $\text{DoRA}+\text{MoDE}_{\mathcal{K}}$. We additionally report the minimal $\text{LoRA}_{\neg\mathcal{K}}+\text{MoDE}_{\mathcal{K}}$ variant, in which MoDE replaces the late-layer LoRA modules entirely and adds only the router and per-layer normalisation ($+0.04\%$ trainable parameters).

\section{Full Commonsense Reasoning Results}
\label{app:commonsense_results}
    Table~\ref{tab:commonsense_reasoning} reports the per-dataset commonsense reasoning and general language-modelling results discussed in \S\ref{exp:commonsense}.
    \begin{table}[h!]
\centering
\caption{
Comparison of MoDE with baselines on five commonsense reasoning datasets and two general language modelling datasets.
$\text{LoRA}_{\neg\mathcal{K}}$ denotes LoRA applied to all transformer layers except the last $|\mathcal{K}|$ late layers, and
$\text{LoRA}_{\text{ALL}}$ denotes LoRA applied to all layers.
Bold and underlined values indicate the best and second-best result within each model block, respectively.
}
\label{tab:commonsense_reasoning}
\begin{sc}
\resizebox{\linewidth}{!}{
\renewcommand{\arraystretch}{1.3}
\begin{tabular}{lccccc cc c}
\toprule
\textbf{Method} & \textbf{ARC-e} & \textbf{ARC-c} & \textbf{BoolQ} & \textbf{OBQA} & \textbf{HellaSwag} & \multicolumn{1}{|c}{\textbf{TruthfulQA}} & \textbf{MMLU} & \multicolumn{1}{|c}{\textbf{Avg.}} \\
\midrule

\textit{LLaMA2-7B} & & & & & & & & \\
\cmidrule(l){1-1}

$\text{LoRA}_{\neg\mathcal{K}}$
& 75.9 & 48.0 & 70.5 & 80.4 & 79.9 & 46.5 & 35.8 & 62.4 \\

$\text{LoRA}_{\text{ALL}}$~\citep{hu2021lora}
& 81.8 & 53.8 & 70.9 & 82.0 & 82.5 & \textbf{49.6} & \textbf{37.3} & 65.4 \\

$\text{DoRA}$~\citep{liu2024doraweightdecomposedlowrankadaptation}
& 85.7 & 57.3 & \underline{72.9} & 83.2 & 88.0 & 49.0 & 37.0 & 67.6 \\

$\text{ReFT}$~\citep{wu2024reft}
& \textbf{87.6} & \textbf{61.3} & 72.2 & 83.3 & \textbf{93.2} & 49.1 & 37.0 & \textbf{69.1} \\

$\text{LoRA}_{\text{ALL}}+\textbf{MoDE}_{\mathcal{K}}$
& 82.2 & 56.4 & 71.4 & \underline{83.4} & 84.0 & \underline{49.3} & \underline{37.1} & 66.2 \\

$\text{DoRA}+\textbf{MoDE}_{\mathcal{K}}$
& \underline{86.0} & \underline{58.1} & \textbf{73.2} & \textbf{83.8} & \underline{88.4} & 49.1 & 37.0 & \underline{67.9} \\

\rowcolor{gray!8}
$\text{LoRA}_{\neg\mathcal{K}}+\textbf{MoDE}_{\mathcal{K}}$ (+0.04\%)
& 82.9 & 53.4 & 71.5 & 82.1 & 83.5 & 48.9 & 36.9 & 65.6 \\

\midrule

\textit{LLaMA3-8B} & & & & & & & & \\
\cmidrule(l){1-1}

$\text{LoRA}_{\neg\mathcal{K}}$
& 83.5 & 70.4 & 70.1 & 78.2 & 90.9 & 51.9 & 53.3 & 71.2 \\

$\text{LoRA}_{\text{ALL}}$~\citep{hu2021lora}
& 84.2 & 71.2 & 70.8 & 79.0 & 91.7 & 52.5 & 54.0 & 71.9 \\

$\text{DoRA}$~\citep{liu2024doraweightdecomposedlowrankadaptation}
& 90.5 & 80.4 & 74.6 & 85.8 & 95.5 & 53.0 & 55.0 & 76.4 \\

$\text{ReFT}$~\citep{wu2024reft}
& \textbf{92.4} & \textbf{81.6} & \textbf{75.1} & \textbf{87.5} & \textbf{96.3} & \textbf{53.2} & \textbf{55.3} & \textbf{77.3} \\

$\text{LoRA}_{\text{ALL}}+\textbf{MoDE}_{\mathcal{K}}$
& 84.8 & 72.0 & 71.3 & 80.0 & 92.2 & 52.3 & 54.2 & 72.4 \\

$\text{DoRA}+\textbf{MoDE}_{\mathcal{K}}$
& \underline{90.8} & \underline{80.8} & \underline{74.9} & \underline{86.2} & \underline{95.8} & \underline{53.1} & \underline{55.2} & \underline{76.7} \\

\rowcolor{gray!8}
$\text{LoRA}_{\neg\mathcal{K}}+\textbf{MoDE}_{\mathcal{K}}$ (+0.04\%)
& 84.6 & 71.5 & 71.0 & 79.5 & 91.9 & 52.2 & 54.0 & 72.1 \\

\bottomrule
\end{tabular}
}
\end{sc}
\end{table}

\section{Additional Analysis}
    

\subsection{Mean and Variance for Routing Pattern Across Tokens}
\label{app:analysis1}
    In this section, we analyse the routing patterns learned with MoDE for the $\mathcal{K}$ ensemble layers during training. We measure the mean and variance of the weights across the training tokens. A higher mean suggests that the model consistently chooses this route, while a higher variance indicates variability in the routes learned for different tokens. We evaluate MoDE trained on top of LoRA and MoDE trained without $k$ LoRA layers using the LLaMA2-7B model on the ARC-e subset.

    According to Figure \ref{fig:meanvar}, for the mean metric, we observe a reverse trend with respect to the sparsity score in Figure \ref{fig:sparsity}. This aligns with our intuition that when the sparsity score of the current route is low, the routing value will be relatively larger than other routes. For the variance, we notice that when MoDE is trained without $k$ LoRA layers, it maintains a high variance throughout tuning. This suggests that many tokens are trained to select this route, but they are dynamically changing. When MoDE is trained with LoRA, both the variance and mean levels stay low, indicating that the other two layers primarily contribute to the final ensemble logits. This suggests that the additional $k$ trainable module within the MoDE framework provides more predictive power to the ensemble layers, aligning with our analysis in \S \ref{sec:sparsity}.

\subsection{MoDE Sparse Routing with Different Top-K Values}
\label{app:topk}
\begin{figure*}[ht!]
        \centering
        \includegraphics[width=\linewidth]{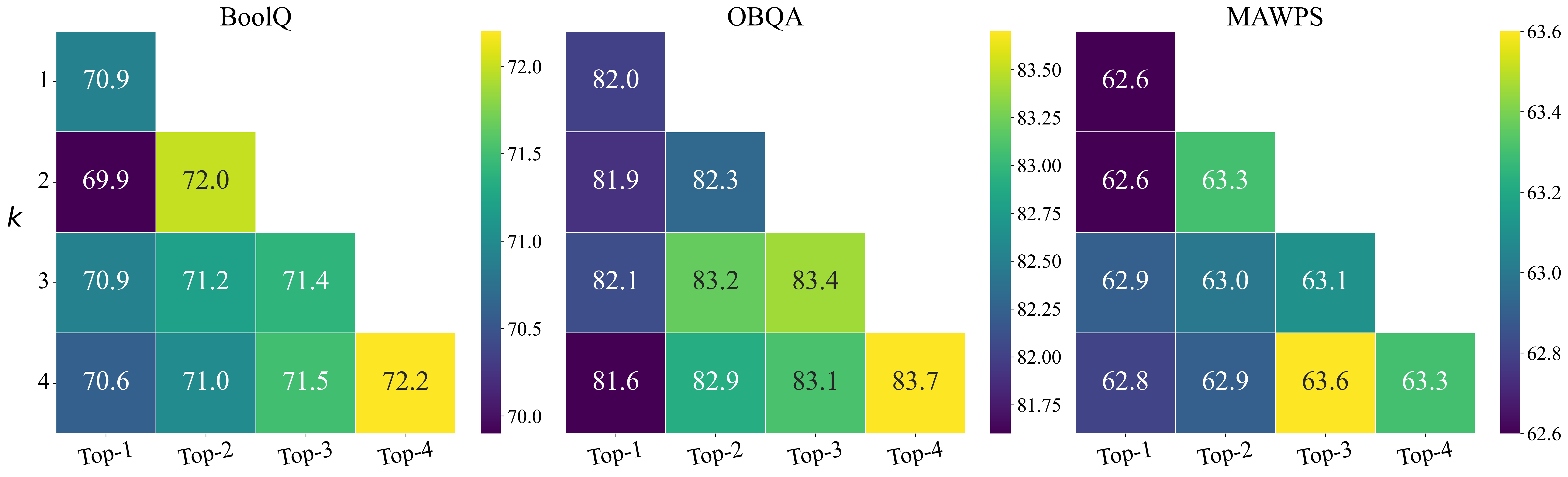} 
        \caption{Accuracy scores for different $k$ ensemble layer ranges and Top-K sparse routing values. Lighter colors indicate better performance. Results evaluated on BoolQ, OBQA, and MAWPS testset.}
        \label{fig:top4}
    \end{figure*}

    We also select a larger ensemble layer range to increase opportunities for early exit. We use $k=4$ for this section, with results for BoolQ, OBQA, and MAWPS presented in Figure \ref{fig:top4}
    
\end{document}